\documentclass[preprint,3p,times,12pt]{elsarticle}

\usepackage{amssymb}
\usepackage[utf8]{inputenc}
\usepackage[T1]{fontenc}
\usepackage{multirow}
\usepackage{xcolor}
\usepackage{newfloat}
\usepackage{hyperref}
\DeclareFloatingEnvironment[name={Figure}]{suppfigure}
\usepackage{ccaption}

\journal{arxiv}

\begin{document}

\begin{frontmatter}

\title{A novel feature-scrambling approach reveals the capacity of convolutional neural networks to learn spatial relations}

\author[1,2,*]{Amr Farahat}
\author[1,3]{Felix Effenberger}
\author[1,2]{Martin Vinck}
\address[1]{Ernst Str{\"u}ngmann Institute for Neuroscience in Cooperation with Max Planck Society, Frankfurt, Germany}
\address[2]{Donders Centre for Neuroscience, Department of Neuroinformatics, Radboud University, Nijmegen, the Netherlands}
\address[3]{Frankfurt Institute for Advanced Studies, Frankfurt, Germany}
\address[*]{amr.farahat@esi-frankfurt.de}

\begin{abstract}
Convolutional neural networks (CNNs) are one of the most successful computer vision systems to solve object recognition. 
Furthermore, CNNs have major applications in understanding the nature of visual representations in the human brain. 
Yet it remains poorly understood how CNNs actually make their decisions, what the nature of their internal representations is, and how their recognition strategies differ from humans. 
Specifically, there is a major debate about the question of whether CNNs primarily rely on surface regularities of objects, or whether they are capable of exploiting the spatial arrangement of features, similar to humans. 
Here, we develop a novel feature-scrambling approach to explicitly test whether CNNs use the spatial arrangement of features (i.e.
object parts) to classify objects. 
We combine this approach with a systematic manipulation of effective receptive field sizes of CNNs as well as minimal recognizable configurations (MIRCs) analysis. 
In contrast to much previous literature, we provide evidence that CNNs are in fact capable of using relatively long-range spatial relationships for object classification. 
Moreover, the extent to which CNNs use spatial relationships depends heavily on the dataset, e.g. texture vs. sketch.  
In fact, CNNs even use different strategies for different classes within heterogeneous datasets (ImageNet), suggesting CNNs have a continuous spectrum of classification strategies. 
Finally, we show that CNNs learn the spatial arrangement of features only up to an intermediate level of granularity, which suggests that intermediate rather than global shape features provide the optimal trade-off between sensitivity and specificity in object classification.  
These results provide novel insights into the nature of CNN representations and the extent to which they rely on the spatial arrangement of features for object classification.  
\end{abstract}


\end{frontmatter}

\section{Introduction}
\label{sec:intro}
The development of Convolutional Neural Networks (CNNs) has led to a revolution in the field of computer vision~\cite{Krizhevsky2012, LeCun2015}.
Machine vision using CNNs has been able to rival human performance in object recognition tasks on large-scale datasets such as ImageNet~\cite{He2016}. 
Moreover, a series of recent works have shown that CNN activations can be used to predict neural activity in the ventral stream of the primate visual system known to be responsible for object recognition~\cite{Yamins2014a, Yamins2016, Cadieu2014a}. 
Therefore, there has been a growing interest in developing behavioral benchmarks that evaluate similarities and differences between CNN models and human vision~\cite{geirhos_general_2018, rajalingham_large-scale_2018, geirhos_partial_2021}.
Crucial to the behavior of these artificial and biological vision systems is their internal representation of objects.
The ability of humans to recognize objects based on their abstract shapes~\cite{landau_importance_1988, biederman_surface_1988, baker2018abstract} suggests that the internal representations of objects in the brain must reflect the global structure of objects~\cite{Biederman1987, barenholtz2006reconsidering}.
An abstract representation of the global shape of an object requires the encoding of the spatial relations between the set of its local features or parts~\cite{Biederman1987, barenholtz2006reconsidering}.
Accordingly, in order to understand the biases that govern the strategies of CNNs performing object recognition, it is central to determine the spatial extent of the diagnostic features CNNs use for object recognition.
Moreover, it is equally important to investigate the role that spatial relations play in the construction of these diagnostic features.

Recent studies have shown inconsistent conclusions regarding the reliance of CNNs trained for object recognition on sets of local features or a global representation of objects~\cite{Jo2017MeasuringTT, Brendel2019, Geirhos2019, cognitive_shape_2017, tartaglini_developmentally-inspired_nodate, baker_deep_2022, Baker2018, kubilius_deep_2016, Baker2020}.
Some studies have shown that CNNs trained for object recognition are biased towards surface statistical regularities (\emph{texture})~\cite{Geirhos2019, Jo2017MeasuringTT, Baker2018, Baker2020, baker_deep_2022}.
In these studies, CNNs were tested on image datasets that included, for example, low-frequency filtered images~\cite{Jo2017MeasuringTT}, shape-texture cue conflict stimuli using style transfer~\cite{Gatys_2016_CVPR, Geirhos2019}, deformed silhouettes and other abstract shape images~\cite{Baker2018, baker_deep_2022} and simple geometric shapes~\cite{Baker2020}.
However, other studies reached different conclusions using other image manipulations or different evaluation methods~\cite{cognitive_shape_2017, tartaglini_developmentally-inspired_nodate, kubilius_deep_2016}.
We reckoned that these different conclusions may be due to the hypothesis-driven approach resulting from the choice of the nature of the stimulus datasets and the object classes represented in them.
For this reason, we developed a hypothesis-free framework for training and testing CNNs that enables us to inspect the shape representations of CNNs by separately controlling the granularity of CNN features (local vs. global) and the spatial relations between them.
This approach allows us to take on the question of to what extent the CNN architecture constrains their capacity to learn shape representations and whether CNNs use the spatial relations among features for object recognition.

Previous work has shown that grid-based image scrambling can be used to identify brain areas sensitive to global configurations of objects~\cite{Grill-spector1998}, expressing characteristic decreases in neural activity with the degree of image scrambling ~\cite{Rainer2002, Grill-spector1998, Vogels1999}. %Feli
Image scrambling, however, disrupts not only the spatial relations between object parts but also the shape of the parts themselves~\cite{margalit2017actually}.
To disentangle these two effects, we developed a feature-scrambling approach that allows us to spatially scramble the pretrained features of CNNs with restricted effective receptive fields (ERFs)~\cite{Brendel2019} without introducing the confounding factors of an image-based scrambling approach. 
The ERF of a CNN is defined as the set of all pixels that can influence the activity of a unit in its last convolutional layer~\cite{le2017receptive}.
These features represent diagnostic parts of the objects at the ERF level of granularity.
After that, we feed these scrambled features to a follow-up CNN that spatially integrates these features and is trained to recognize the class of objects.
Recent work suggests that CNNs with restricted ERF sizes can achieve a performance similar to regular CNNs on ImageNet~\cite{Brendel2019}. 
However, it remains unclear whether these models use the same strategies as regular CNNs to solve the task.
Notably, the approximation of regular CNNs performance on ImageNet with CNNs with restricted ERFs implies that CNNs rely on a classification strategy that pools local evidence from separate locations in the image without learning the spatial relations between them.
This observation would predict, for instance, that training a follow-up CNN on the pretrained features of a CNN with restricted ERFs should minimally affect performance.
It would also predict that spatially scrambling the pretrained input features to the follow-up CNN would not lead to a significant difference in performance to training with the right spatial arrangement of the features.
In this work, we tested these predictions on different datasets that comprise texture-rich and texture-less images to examine whether CNNs employ different classification strategies for different datasets.
Furthermore, we examined to what extent CNNs with smaller ERFs develop representations similar to CNNs with larger ERFs. 
Finally, we performed a minimal recognizable configuration (MIRC) analysis~\cite{Ullman2016} to quantify the minimal image patch sizes required by CNNs to achieve correct classification.

\section{Methods}
\label{meth}
\subsection{Datasets}

We trained CNNs on three datasets with different feature characteristics: the Sketchy, Animals, and ImageNet datasets. 
The Sketchy dataset contains 75,471 human-drawn sketches spanning 125 classes~\cite{sketchy2016}. 
Each sketch is a textureless, black-and-white bitmap graphic that only contains information about the contours of objects without any surface proprieties, and sketches have a high degree of intra-class variability (Fig.~\ref{fig:fig1}c).
The Animals dataset consists of 37,322 color images spanning 50 classes~\cite{animals} (Fig.~\ref{fig:fig1}b).
The well-known ImageNet dataset contains 1.2M color images across 1000 classes~\cite{deng2009imagenet} that span different animals and man-made artifacts.

\subsection{Models}
\label{sec:models}

We created residual CNNs~\cite{He2016} with ERFs of variable sizes (Table~\ref{table:1}) by changing the size of the filters of different residual units across layers~\cite{Brendel2019}. 
The residual CNNs consist of $4$ blocks that contain $2$, $3$, $3$, and $2$ residual units, respectively. 
Each residual unit consists of $3$ convolutional layers: The first and last layers always have filters of size $1\times1$ and the filter size of the middle layer varies according to Table~\ref{table:1}. 
Adjusting the filter size of the residual units results in models with  ERFs of either $11$, $23$, $47$, $95$, or $227$ pixels squared in the last layer. 
We refer to these models by their ERF sizes, writing ERF23 for a network with an ERF of size $23\times 23$ pixels. 
Note that since our input images are always of size $224\times224$ pixels, only the model ERF227 has units in the last convolutional layer with ERFs covering the entire image, before features are globally averaged across spatial locations in the penultimate layer.

\begin{table*}[]
\centering
\begin{tabular}{ccccccccc}
\hline
\multirow{2}{*}{\textbf{Blocks}} & \multirow{2}{*}{\textbf{\begin{tabular}[c]{@{}c@{}}Residual\\ Units\end{tabular}}} & \multirow{2}{*}{\textbf{\begin{tabular}[c]{@{}c@{}}Feature\\ Maps\end{tabular}}} & \multicolumn{1}{l}{\multirow{2}{*}{\textbf{Stride}}} & \multicolumn{5}{c}{\textbf{Filter Sizes}}                                                                                                                                               \\ \cline{5-9} 
                                 &                                                                                    &                                                                                  & \multicolumn{1}{l}{}                                 & \multicolumn{1}{l}{\textbf{ERF11}} & \multicolumn{1}{l}{\textbf{ERF23}} & \multicolumn{1}{l}{\textbf{ERF47}} & \multicolumn{1}{l}{\textbf{ERF95}} & \multicolumn{1}{l}{\textbf{ERF227}} \\ \hline
Block 1                          & 2                                                                                  & 128                                                                              & 2                                                    & 3,3                                & 3,5                                & 3,5                                & 3,5                                & 5,5                                 \\
Block 2                          & 3                                                                                  & 256                                                                              & 2                                                    & 1,1,1                              & 3,1,1                              & 3,3,5                              & 3,3,5                              & 5,5,5                               \\
Block 3                          & 3                                                                                  & 512                                                                              & 2                                                    & 1,1,1                              & 1,1,1                              & 1,1,1                              & 3,3,3                              & 5,5,5                               \\
Block 4                          & 2                                                                                  & 1024                                                                             & 1                                                    & 1,1                                & 1,1                                & 1,1                                & 1,1                                & 5,5                                 \\ \hline
\end{tabular}
\caption{Architecture details for our ResNets of different ERFs.}
\label{table:1}
\end{table*}

\subsection{Feature-scrambling approach}

For the feature-scrambling approach, we build CNN models that are composed of two sub-networks, a \emph{base network} and a \emph{follow-up network}~(Fig.~\ref{fig:fig1}d). 
The base network transforms the image to high-level feature maps of a given size by being trained on image classification in a standalone way. 
These pretrained features are then fed into a follow-up network. 
The follow-up network then
further transforms these feature maps in a series of convolutional layers.
Finally, features are pooled in a location-discarding way in a global average pooling layer and then a Softmax classification layer.
This approach allows us to independently examine the granularity of features used by CNNs for object recognition and to determine to what extent the spatial relations among them contribute to their performance. 

We used networks with different ERFs as base networks. 
We trained them separately for image classification and then detached the fully connected classification layer and the global average pooling layer of the trained network and used it with frozen weights as the base network in our feature-scrambling approach. 
Subsequently, we attached the follow-up network such that it receives the features of the pretrained base networks as inputs in either a scrambled or unscrambled way. 
Specifically, for the unscrambled case, we passed the feature maps unchanged to the follow-up network.
For the scrambled case, we generated random indices once and used them to permute the feature vectors across spatial locations.
The follow-up network is a residual block formed of four residual units. 
We differentiated between two types of follow-up networks: \textit{with} or \textit{without} spatial aggregation:
(1) A follow-up network \textit{with} spatial aggregation has a stride of $2$ for its first two residual units and filter size $3\times3$ for all its residual units.
(2) A follow-up network \textit{without} spatial aggregation is formed exclusively of convolutional layers with filter size $1\times1$ and no down-sampling. 
In summary, for each of our base networks (ERF11, ERF23, ERF47, ERF95, and ERF227), we trained 3 models depending upon: (1) the type of the follow-up network (with or without spatial aggregation); (2) scrambling the features between the two sub-networks or not.

The considered models can be summarized as follows:
\begin{itemize}
\setlength\itemsep{0.1em}
    \item \textbf{Base:} only the base network trained in a standalone way.
    \item \textbf{Base + Follow-up without scrambling:} the model is formed of the pretrained base network plus the follow-up network with spatial aggregation and without feature-scrambling. 
    \item \textbf{Base + $1\times1$ Follow-up without scrambling:} the model is formed of the pretrained base network plus the follow-up network without spatial aggregation and without feature-scrambling. 
    This model serves as a control for the significance of increasing the ERF of the model by adding the follow-up network. 
    \item \textbf{Base + Follow-up with scrambling:} the model is formed of the pretrained base network and the follow-up network with spatial aggregation and with global feature-scrambling during training. 
\end{itemize}

Additionally, the \textbf{Base + Follow-up without scrambling} models were tested while the input features to the follow-up network were randomly scrambled either globally or locally. 
\subsubsection{Training}

All simulations were performed using the TensorFlow library~\cite{tensorflow2015-whitepaper}.
We used stochastic gradient descent with momentum $=0.9$ to update the weights with initial learning rate $=0.01$ for the first $10$ epochs followed by exponential decay for the rest of training.
For the ImageNet dataset, we trained for $50$ epochs, and for the animals and Sketchy datasets, we trained for $75$ epochs.

During training, for non-square images, we first cropped the central square portion of the image with the shortest dimension of the image to keep the aspect ratio of the objects constant before resizing the images to $256\times256$ pixels.
We then applied minimal data augmentation in the form of random right and left horizontal flipping of the images, followed by random cropping of $224\times224$-pixel patches used for training.
During testing, after centrally cropping the images, we resized them to $256\times256$ pixels, and then we cropped the central $224\times224$-pixel patch.

\subsection{Representational similarity analysis (RSA)}
We used RSA to investigate the representations of the CNNs of different ERFs~\cite{Nili2014}. 
To avoid the results being biased to the number of classes in each dataset, we sampled 50 random classes from each dataset (the lowest number of classes in the three datasets).
Then we sampled $8$ random images from each class for a total of $400$ images, ran them through all the models of different ERFs, and extracted the activations of the last convolutional layer of each residual unit ($n=10$), the global average pooling layer (GAP) and the Softmax layer.
For the Sketchy and Animals datasets, we averaged the layers' RDMs across 5 repetitions of random initialization.
We created the representation dissimilarity matrix (RDM) for each layer by computing the pairwise correlation distance for its activations ($400\times400$ matrix).
Next, we computed a second-order RDM for all the layers of the models ($60\times60$ matrix) by computing the correlation distance between the upper triangle of the layers' RDMs.
For visualization purposes, we used multi-dimensional scaling (MDS) to reduce the dimensionality of the second-order RDM to two dimensions.

\subsection{Minimal recognizable configurations analysis (MIRC)}

We adopted the MIRC analysis~\cite{Ullman2016} previously used for humans for CNNs. 
MIRC analysis is a recursive process that looks for the smallest image patches that still yield a correct classification result.
MIRC analysis starts with a given, correctly classified image of class $c$. 
Starting from the whole image as one patch, four descendant patches are created recursively from each patch, with each descendant spanning 75\% of the height and width of the patch at the previous level. 
Each patch is then upsampled using bilinear interpolation to $224\times 224$ pixels to match the input size of the models.
The recursive subdivision process continues for each patch as long as the patch is still correctly classified as belonging to class $c$. 
Subdivision stops once the classification of a patch is no longer correct. 
This process defines a tree structure and the leaves of the tree are the MIRCs.
The level of a leaf node in the tree is referred to as the level of the MIRC it represents. 
By construction, the higher the MIRC level, the smaller the patch of the image used for classification.

\section{Results}

\subsection{Feature scrambling during training and testing}

\begin{figure*}[h!]
    \centering
    \includegraphics[width=\linewidth]{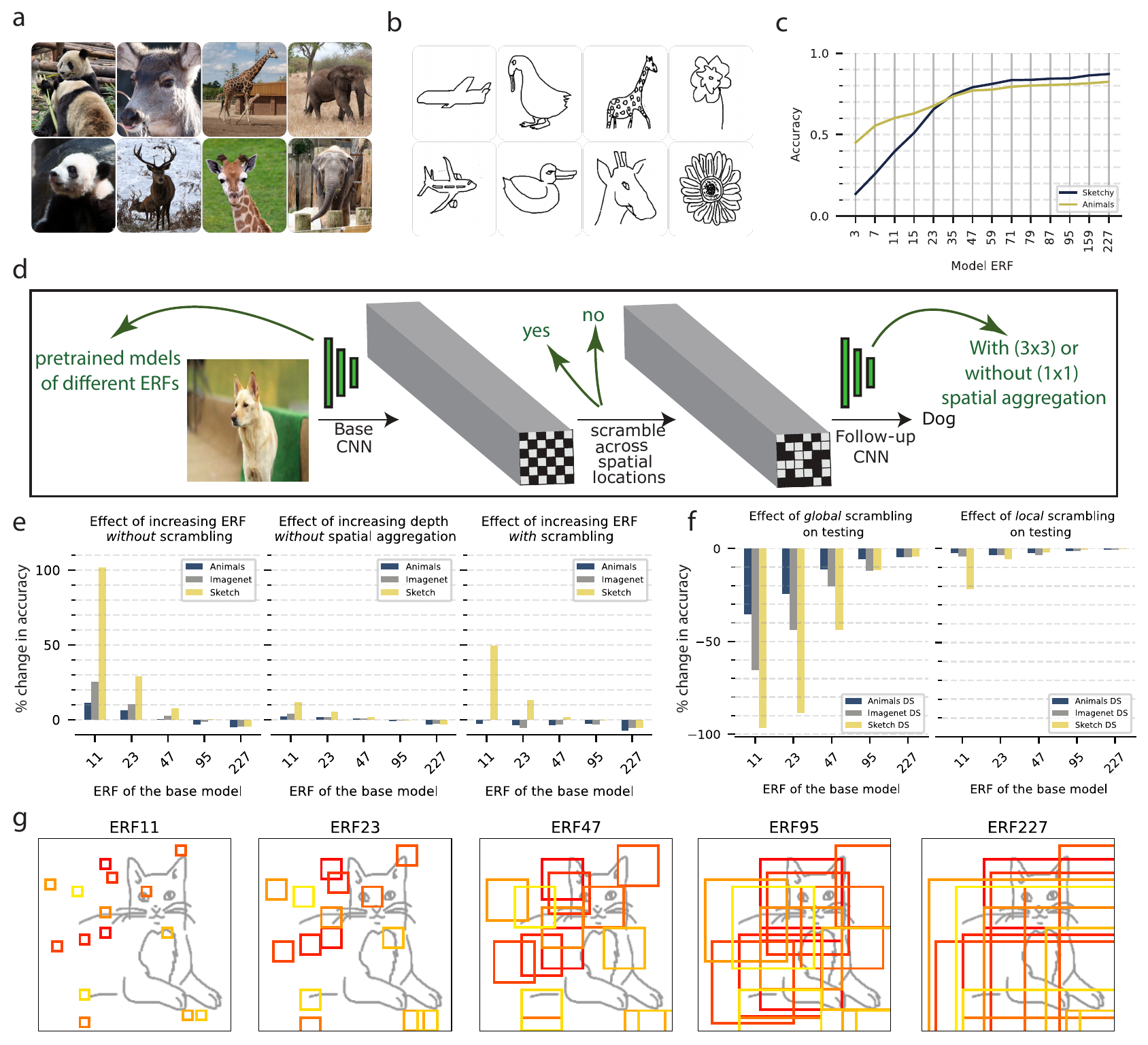}
    \caption{
    \textbf{Feature scrambling during training and testing.} 
    \textbf{(a, b)} Example images for the Animals and Sketchy datasets, respectively.
    \textbf{(c)} CNN performance as a function of the ERF, separately for the Sketchy and Animals datasets.
    \textbf{(d)} A schematic for the feature-scrambling approach.
    \textbf{(e)} Effects of adding the follow-up network to the pretrained base networks either with spatial aggregation without scrambling (left), with spatial aggregation with scrambling (right), or without spatial aggregation (middle).
    \textbf{(f)} Effect of global and local feature-scrambling on the testing performance of the base + follow-up models with spatial aggregation without scrambling.
    \textbf{(g)} A schematic depicting the ERF of random artificial neurons in the last convolutional layer of models of different ERFs.
    }
    \textbf{}
    \label{fig:fig1}
\end{figure*}

We trained CNNs of different ERF sizes on three different datasets: the Sketchy~\cite{sketchy2016}, the Animals~\cite{animals} (Fig.~\ref{fig:fig1}a-b), and the ImageNet~\cite{deng2009imagenet} dataset. 
Example ERFs of five models are shown in Fig.~\ref{fig:fig1}g.
We note that the ERF is a theoretical upper limit on the set of pixels that can activate a given unit, and that not all pixels of the ERF necessarily activate the corresponding deep unit, depending on connection weights.
We found that CNN performance increased with ERF size for both the Sketchy and Animal datasets, with a visible saturation for larger ERFs. 
However, CNN performance depended more strongly on the ERF size for the Sketchy dataset than for the Animals dataset (Fig.~\ref{fig:fig1}c).
Because changing the filter sizes across models will also induce changes in the number of trainable parameters in the models and consequentially their expressive capacity, we performed a control experiment in which we created wide networks with small ERFs but matched the number of parameters of the network with the largest ERF (ERF227).
We found a slight increase in accuracy, but the models still showed a substantial reduction in performance compared to the corresponding network with a large ERF (Fig. A.\ref{fig:control}).

\begin{figure*}[h!]
    \centering
    \includegraphics[width=\linewidth]{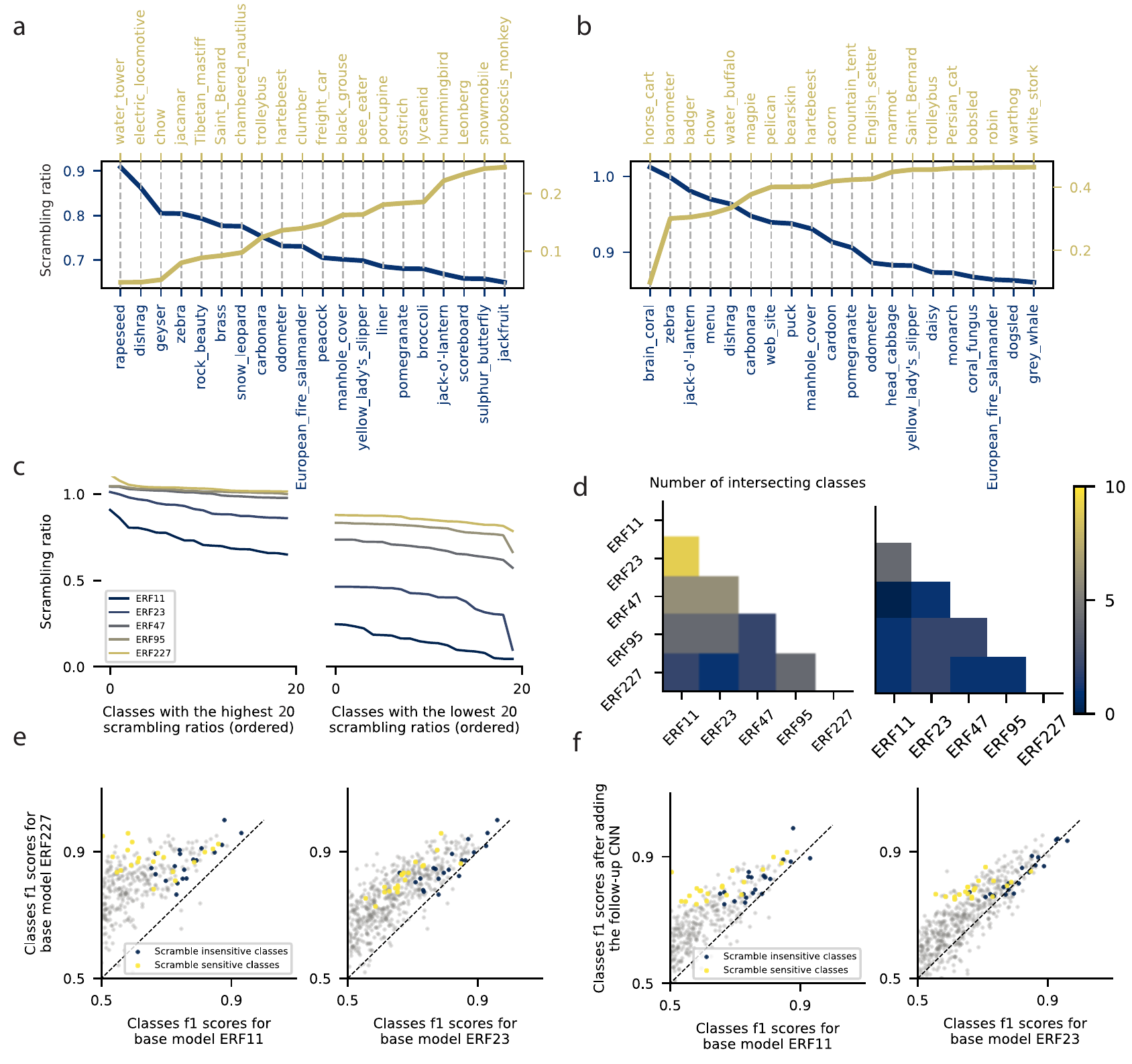}
    \caption{
    \textbf{(a, b)} The 20 least (in blue) and most (in yellow) scrambling-sensitive ImageNet classes for the models ERF11 (a) and ERF23 (b).
    \textbf{(c)} Scrambling ratios of the 20 least (left) and most (right) scrambling-sensitive ImageNet classes for models of different ERFs. 
    High and low values of the scrambling ratio indicate that feature-scrambling has minor and major effects on class performance, respectively. 
    \textbf{(d)} Number of the intersecting classes for the 20 least (left) and most (right) scrambling-sensitive ImageNet classes among models of different ERFs.
    \textbf{(e)} $f1$ performance scores of ImageNet classes for ERF11 and ERF23 models against ERF227 model. In blue and yellow are respectively the least and most scrambling-sensitive classes. 
    \textbf{(f)} $f1$ performance scores of ImageNet classes for the base model vs. base model after adding the follow-up network. In blue and yellow are respectively the least and most scrambling-sensitive classes. 
    } 
    \label{fig:fig2}
\end{figure*}

The dependence of the classification performance on ERF size suggests that the network's ERF has a major impact on object recognition, especially for textureless datasets such as the Sketchy dataset. 
One explanation for the observed performance increase could be that CNNs with large ERFs can learn to exploit relatively large-scale features, which are especially important for texture-less datasets. 
However, the comparison between networks with large ERFs and small ERFs does not yet provide direct evidence that CNNs with large ERFs rely on large-scale shape features. 
For example, it is possible that the pooling in large ERFs does not take into account the spatial configuration among the features. 
Instead, the network might just accumulate local evidence in a different manner than networks with smaller ERFs. 
This reasoning suggests that in order to investigate the network's sensitivity to the spatial configuration of features, it is necessary to distort (i.e. scramble) the spatial arrangement of features and then test the impact of this distortion. 
Importantly, this scrambling should be done at the level of the network features rather than at the image level, as the latter often leads to confounding high-contrast image artifacts. 
Specifically, we took the following approach: 

1) We trained a network with a small ERF size on an object recognition task. 
We call this the base CNN, which was not further modified.% 

2) We then trained a follow-up network, which received input from the last convolutional layer of the pretrained base CNN. 
These pretrained input features represent diagnostic features of certain granularity depending on the ERF of the base CNN i.e. object parts at different scales.
The follow-up network has an ERF that covers the entire image. 
We observed that adding the follow-up network led to an increase in performance compared to the base network. 
Consistent with the ERF survey experiment (Fig.~\ref{fig:fig1}c), the increase in performance was relatively small for the ImageNet and Animals datasets but was large for the Sketchy dataset for base networks with smaller ERFs (Fig.~\ref{fig:fig1}e, left  panel). 
Absolute performances are shown in Fig. A.\ref{fig:rawdata}.

3) To rule out the possibility that the observed performance increase for such stacked networks was just caused by increasing the depth of the model by appending the follow-up network, we trained a follow-up network that consisted only of $1\times1$ convolutions without strides to prevent spatial aggregation. 
We observed only a slight increase in accuracy for all datasets (Fig.~\ref{fig:fig1}e middle), which shows that spatial aggregation of inputs was crucial for the observed performance boost (Fig.~\ref{fig:fig1}e left).

4) To examine whether the spatial configuration of features mattered, we trained the same follow-up networks after spatially scrambling the features in the last convolutional layer of the base network. 
We used a fixed spatial permutation (i.e. scrambling) of these features that was constant during training. 
We observed a smaller increase in performance for the Sketchy dataset for ERFs 11 and 23 (Fig.~\ref{fig:fig1}e right). 
Furthermore,  no further increase in accuracy could be observed for the Animals and ImageNet datasets in this case (Fig.~\ref{fig:fig1}e right). 
Taken together, these findings suggest that CNNs can learn to utilize the configuration of spatially distant features when constructing more complex features in subsequent layers, especially for datasets in which shape is expected to be critical for object classification. 

\begin{figure*}
    \includegraphics[width=\linewidth]{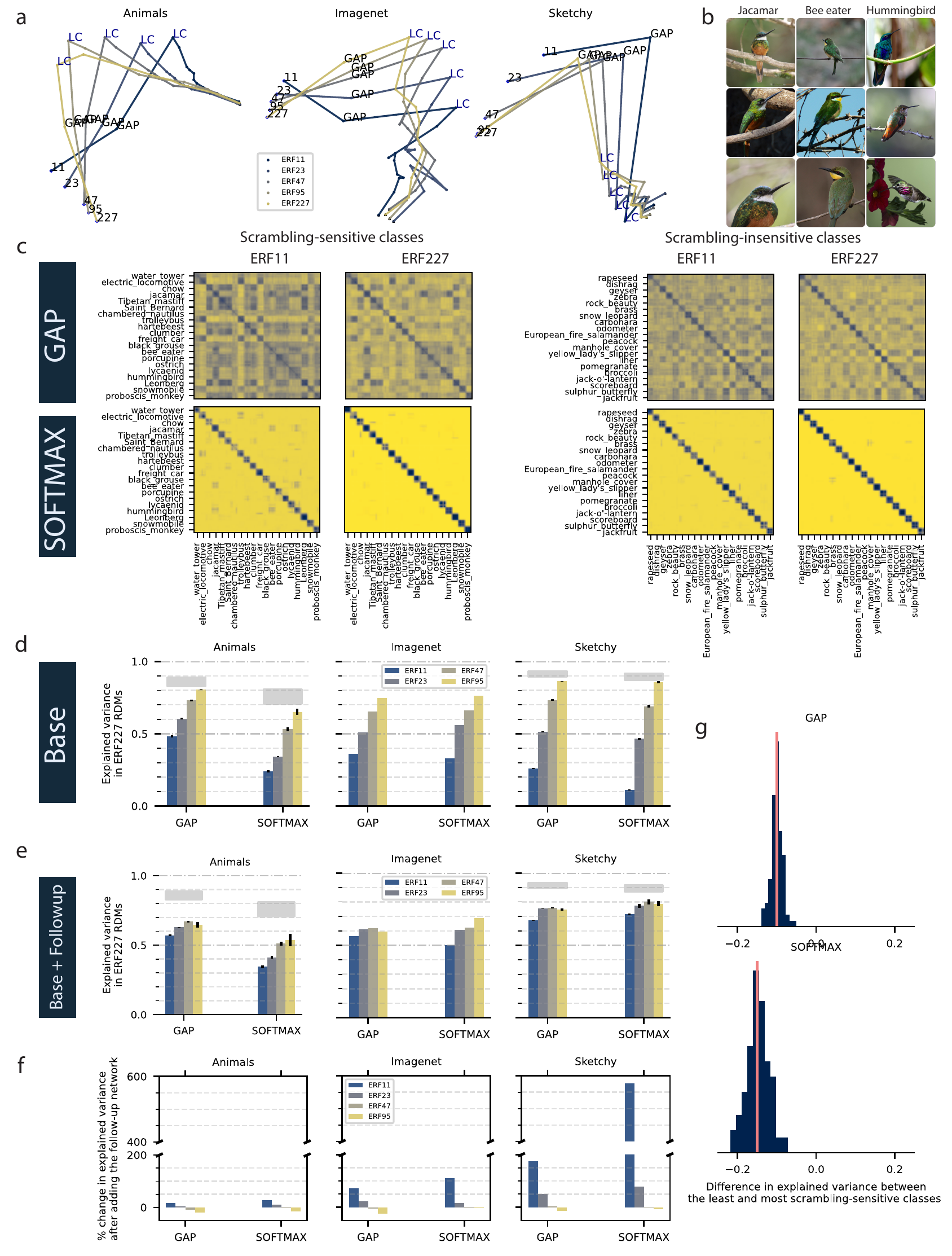}
    \caption{See next page
    }
    \label{fig:fig3}
\end{figure*}
        
\begin{figure*}[t]
    \centering
    \contcaption{
    \textbf{(a)} Representation trajectories for five CNNs with different ERFs trained on 3 different datasets. For the Sketchy and Animals datasets, we averaged the layers' RDMs (Representational Dissimilarity Matrices) of 5 training iterations of each model of a certain ERF size before computing the second-order RDM of all layers.
    \textbf{(b)} Each column shows three examples from the ImageNet dataset for three bird classes.
    \textbf{(c)} RDMs of the global average pooling (GAP) and Softmax layers for the models ERF11 and ERF227 computed separately on the 20 least and most feature-scrambling sensitive ImageNet classes as estimated using the ERF11 model and the feature-scrambling approach. We sampled 20 images randomly from each class so each RDM is $400\times400$ (better viewed digitally).
    \textbf{(d)} The amount of explained variance ($R^{2}$) by the GAP and Softmax layers' RDMs of models with different restricted ERFs in the RDMs of the ERF227 model.
    \textbf{(e)} The amount of explained variance ($R^{2}$) by the GAP and Softmax layers' RDMs of models with different restricted ERFs after adding the follow-up network in the RDMs of the ERF227 model.
    \textbf{(f)} Percentage change in the amount of explained variance by RDMs of models with different restricted ERFs in the RDMs of the ERF227 model after adding the follow-up network that increases the ERF of the models to cover the whole image.
    \textbf{(g)} 
    The distributions of the difference in explained variance by ERF11 model RDMs in ERF227 model RDMs between scrambling-sensitive and scrambling-insensitive classes.
    RDMs were computed by randomly sampling images separately from the scrambling-sensitive and scrambling-insensitive classes.
    The number of repetitions is 100.
    }
\end{figure*}

5) As a complementary approach to the fixed scrambling during training, we also performed random feature scrambling during testing.
As before, the scrambling was again done at the last convolutional layer of the base network. 
As predicted, we observed a general decrease in the accuracy of the models with spatial aggregation (base + follow-up) when the features were globally scrambled during testing (Fig.~\ref{fig:fig1}f left). 
This effect depended strongly on the dataset, with a relatively weak effect for the Animals dataset and a very strong effect for the Sketchy dataset. 
Moreover, the performance reduction was particularly pronounced for models with small ERFs that exclusively encode local features of fine granularity before the scrambling is done.
It is worth noting that this effect cannot be simply explained by the type of the dataset (sketches versus natural images) since the reduction in performance varied substantially between the Animals and ImageNet datasets, even though both consist of natural images.

6) As a control, we also performed a ``local'' scrambling, in which the features were scrambled only at neighboring locations. 
The reduction in performance with local scrambling was much weaker compared to global scrambling, indicating that the loss of performance with global scrambling is due to the distortion of the global configuration of the features, not the confounding effects of the scrambling process itself.

Together, these results highlight the importance of the granularity of features and their spatial configurations for object recognition, especially for datasets in which texture is less informative. 
In other words, models with larger ERF can extract more coarse-grained features, which are more diagnostic for the object class, i.e. have higher accuracy and are less susceptible to scrambling.
These coarse-grained features are diagnostic on their own and do not need to be spatially integrated to construct more complex features in subsequent layers (the follow-up network).
However, the granularity of these features differs between datasets.

\subsection{Variability of classification strategies between classes in ImageNet}
%
% Felix:
Depending on the dataset, we observed different effects of ERF sizes and feature scrambling on network classification performance.
Changing the ERF size had the weakest effect on performance for the Animals dataset and the strongest effect for the sketches dataset, with ImageNet in between (Fig.~\ref{fig:fig1}e left). 
The strongest effect of feature scrambling was observed on the Sketchy dataset, followed by ImageNet and then the Animals dataset, which was least affected by feature scrambling (Fig.~\ref{fig:fig1}f left).
These findings can be explained by the image statistics in the different datasets.
Two extremes are given by the Animals and Sketchy dataset: While images in the Animals dataset can already be classified using local textural features, pictures in the Sketchy dataset require the integration of spatially distant features for classifications.
For ImageNet, the classification may allow for different class-specific strategies (e.g., animals vs. man-made artifacts). 
To test the hypothesis that CNNs use different classification strategies for different ImageNet classes, we used the feature-scrambling approach described above.  
As a measure of how CNN classification performance is affected by global feature scrambling, we consider the scrambling ratio as the ratio of class $f1$ scores before and after scrambling.
A high scrambling ratio indicates that a class is not sensitive to feature scrambling (which we call scrambling-insensitive), and a low value indicates sensitivity to scrambling (which we call scrambling-sensitive).
This ranks the classes according to their sensitivity to the global spatial feature configuration in the last CNN layer of the base network~(Fig.~\ref{fig:fig2}a-c). 
For this analysis, we only considered classes that the model reliably classified before scrambling ($f1 > 0.75$).

As hypothesized, the least scrambling-sensitive classes predominantly express characteristic surface patterns (texture) such as the rapeseed, brain coral, and zebra classes (Fig.~\ref{fig:fig2}a and b in blue for base models ERF11 and ERF23 respectively).
Scrambling-sensitive classes, on the other hand, were not found to express such characteristics textures, such as the water tower, electric locomotive, and horse cart classes (Fig.~\ref{fig:fig2}a and b in yellow for base models ERF11 and ERF23 respectively). 
We hypothesized that the variability in scrambling sensitivity was due to the intrinsic properties of the classes and their performance at low ERFs, rather than to the scrambling operation itself. 
In fact, we found that classes with high scrambling sensitivity only exhibited this high sensitivity for models with small ERFs (Fig.~\ref{fig:fig2}c right). 
However, the scrambling sensitivity of classes was found to be mostly independent of ERF size~(Fig.~\ref{fig:fig2}c left). 
To confirm that this effect is a consequence of the heterogeneity of the ImageNet dataset and not the ordering process, we repeated the same analysis for the Animals dataset and did not observe such substantial variability in the scrambling ratios among classes, e.g., for the base model ERF11, scrambling ratios ranged from $0.05$ to $0.91$ and from $0.61$ to $0.96$ for ImageNet and Animals datasets respectively. 
We furthermore found that the set of the least scrambling-sensitive classes is mostly consistent across models (Fig.~\ref{fig:fig2}d left).
This is in contrast to the set of the most scrambling-sensitive classes (Fig.~\ref{fig:fig2}d right). 
Thus, the performance of scrambling-sensitive classes depends more on the models' ERFs and, therefore, relies on features of coarser granularity. 

Therefore, we hypothesized that the scrambling ratio should predict the performance increase from the ERF11 to the ERF227 network (Fig. A.\ref{fig:rawdata}), as well as the performance increase obtained by adding the follow-up network to the pretrained base network (Fig. A.\ref{fig:rawdata} and Fig.~\ref{fig:fig1}e). 
Indeed, the performance increase for ERF227 compared to ERF11 and ERF23 was greater for scrambling-sensitive than for scrambling-insensitive classes (Fig.~\ref{fig:fig2}e). 
Specifically, the performance ($f1$ score) of the model ERF227 on the 20 most scrambling-sensitive classes was higher than that of all other models (ERF11, ERF23, ERF47, and ERF95) in a statistically significant way according to the Wilcoxon signed-rank test.
In contrast, for the 20 least scrambling-sensitive classes, the performance of the ERF227 model was only significantly higher than the models ERF11, ERF23, and ERF47, but not ERF95. 
Similarly, the performance increase caused by the addition of a follow-up network was larger for scrambling-sensitive classes than for scrambling-insensitive classes (Fig.~\ref{fig:fig2}f). 
In particular, increasing the ERF of the models by adding the follow-up network led to a statistically significant increase in the performance of the 20 most scrambling-sensitive classes for the models ERF11, ERF23, ERF47, and ERF95.
For the 20 least scrambling-sensitive classes, it only led to a statistically significant increase in performance for the models ERF11 and ERF23.

\begin{figure*}
    \centering
    \includegraphics[width=\linewidth]{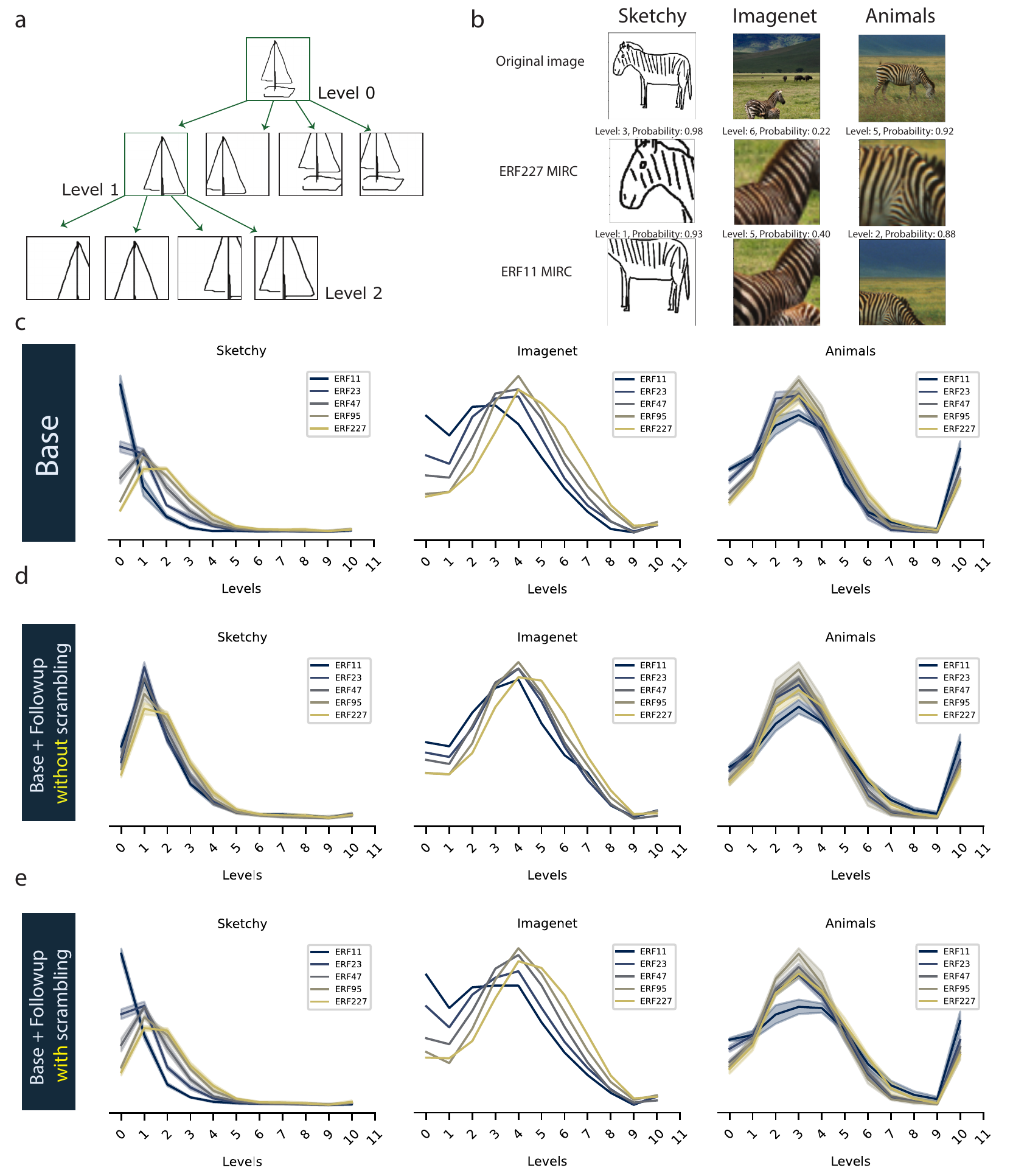}
    \caption{
    \textbf{(a)} Illustration of the MIRC procedure. 
    \textbf{(b)} Example MIRCs for three different images (first row) of the zebra class from three datasets (Sketchy, ImageNet, and Animals) for the models ERF227 (second row) and model ERF11 (third row). The MIRCs shown are the MIRCs with the highest probability among the MIRCs of the highest level of that image.
    \textbf{(c, d, e)} Distribution of the maximum MIRC level for each correctly classified image in the test dataset for the Sketchy, ImageNet, and Animals datasets, respectively for the base networks of different ERF sizes (c), after adding the follow-up network without scrambling (d), and after adding the follow-up network with the spatial scrambling of its input features (e). 
    For the Sketchy and Animals datasets, the histograms are averaged over 5 training iterations. The shaded area represents the standard deviation
    The high frequency of images with MIRCs of level 10 in the Animals dataset is because of the images that belong to the classes that the models usually predict when the correct class cannot be identified. 
    }
    \label{fig:fig4}
\end{figure*}
\subsection{Representation Similarity Analysis}
\label{RSA}

Next, we investigated the role of ERF size on the classification strategies used by CNNs.
We used representation similarity analysis (RSA)~\cite{Nili2014} to test whether CNNs of different ERF sizes develop comparable representations, reflecting similar or different classification strategies (Fig.~\ref{fig:fig3}a-c; see section \ref{meth}).
For each layer, we computed a representation dissimilarity matrix (RDM) by computing the pairwise correlation distance on the activations resulting from different images. 
% % % 
We then computed the dissimilarity (using the pairwise correlation distance) of the RDMs between all layers of all models, thus indicating the similarity of the representations between different layers of different models. 
We observed that models with comparable ERFs are closer in the low dimensional space (Fig.~\ref{fig:fig3}a), indicating that the distances among the corresponding layers of the models depend on the models' ERF. 

To further investigate whether CNNs with small ERFs use classification strategies similar to those of standard CNNs with large ERFs, we correlated the RDMs for all models with the RDM for the ERF 227 model. 
Specifically, we computed the variance explained ($R^{2}$) between the RDMs of the ERF227 model and the RDMs of the models with smaller ERFs (Fig.~\ref{fig:fig3}d). 
This was done separately for the Global Average Pooling (GAP) and Softmax layers for the three datasets. 
For both GAP and Softmax, we observed a gradual increase in the amount of explained variance with ERF size, i.e., models with small ERFs are more dissimilar to the ERF227 model. 
The amount of explained variance depended on the dataset: 
The Sketchy dataset had the lowest amount of explained variance for models with small ERF, followed by ImageNet and the Animals datasets. 
This result agrees with the differences between datasets in terms of the models' classification performance (Fig.~\ref{fig:fig1}e).
We repeated the same analysis after adding the follow-up networks to the base models, which in each case increased the ERF to cover the whole image (e.g.\ $235 pixels^{2}$ for ERF11 base model) (Fig.~\ref{fig:fig3}e).
We noticed an increase in the amount of explained variance after adding the follow-up network, especially on the Sketchy dataset and for the models with small ERFs (Fig.~\ref{fig:fig3}f).
Again, there was only a minor and intermediate increase for the animals and ImageNet databases, respectively. 
This supports the notion that CNNs can deploy different classification strategies depending on their ERF.

Moreover, CNN classification strategies should also differ among object classes even within the same model.
In this regard, we hypothesized that the explained variance between ERF11 and ERF227 should differ between the scrambling-sensitive and scrambling-insensitive classes. 
In particular, we expected that the explained variance should be smaller for scrambling-sensitive classes because for those classes, one expects spatial integration. 
For that purpose, we selected the 20 most and least scrambling-sensitive classes of the ImageNet dataset as determined by our feature-scrambling approach for the model ERF11 (Fig.~\ref{fig:fig2}a), randomly selected 20 images from each class, passed them through the models ERF11 and ERF227, computed the RDMs of the GAP and Softmax layers for each condition separately. 
We repeated the process 100 times to create a distribution of the difference in the variance explained by model ERF11 in model ERF227 between the scrambling-sensitive and scrambling-insensitive classes. 
Indeed, we observed the expected difference (Fig.~\ref{fig:fig3}g). 
Additionally, by visual inspection, the difference between the RDMs of the models ERF11 and ERF227 calculated on the scrambling-sensitive classes is especially pronounced in the off-diagonal part of the matrix, which represents the similarity among the inter-class pairs of images (Fig.~\ref{fig:fig3}c left two columns). 
We hypothesized that the reason behind this difference is that the ERF11 model extracts lower-level features that are not indicative of a specific class, but rather shared among multiple classes.
For example, we observe these blocks of low dissimilarity in Fig.~\ref{fig:fig3}c (most lower left panel) between the class jacamar and the classes bee-eater and hummingbird, which have shared color and local features (Fig.~\ref{fig:fig3}b). 
Together, these results further support our conclusion that the granularity of features used by CNNs (which in terms are determined by their ERF sizes) plays a crucial role in their ability to perform object recognition.
Moreover, the granularity of the CNN features is determined not only by its ERF but also by the statistics of the images in the datasets, separately for each class.
Although more coarse-grained features can be more reliable for object recognition, they are only exploited by CNNs when needed e.g. the Sketchy dataset and scrambling-sensitive classes in ImageNet.
This agrees with the simplicity bias in CNNs (and more generally all neural networks) when trained with a gradient-based learning rule: %
Networks tend to become selective to the easiest (and most local) features that allows them to solve the classification task at hand.

\subsection{Minimal recognizable configurations (MIRCs) analysis}
The results so far suggest that CNNs recognize objects based on features that vary in their granularity depending on the dataset and the object class. 
For datasets and object classes that have relatively little or no texture information, CNNs can learn to construct diagnostic features of coarser granularity from more fine-grained features by exploiting the spatial relations between them. 
This raises the following questions: 1) What is the spatial extent of these coarse features and spatial relations learned by CNNs?
2) What is the advantage of more coarse-grained features over more fine-grained features for object recognition? 
The feature-scrambling results shown above indicate that even for the Sketchy dataset, increasing the ERF of the base models beyond $47\times47$ had a limited effect on performance. 
This result suggests that the features required for reliably recognizing objects are still predominantly local, i.e., they span maximally about 4-5\% of the image. 

To further test the reliability of the features utilized by models of different ERFs and visualize them in the image space, we performed a MIRC analysis. 
MIRC analysis tests the ability of the models to categorize images based on localized image patches by searching for the minimal (i.e. smallest) feature configurations in the image that are still correctly recognizable by the models. 
We searched for the MIRCs of each image in the test dataset of the Sketchy and Animals datasets, and randomly sampled one-third of the images in the test dataset of the ImageNet dataset.
For each image, we cropped $75\%$ of the image starting from each corner so that each image yields 4 descendants (Fig.~\ref{fig:fig4}a). 
We then upsampled each descendent crop to the original image size ($224\times224$) and used the model to predict its object class. 
We repeated the process for each descendant that was correctly classified by the model until we reached the image that was correctly identified by the model but had no correctly classified descendants.
This image was declared a MIRC and its level in the search tree defines its size, i.e.\ the deeper (higher) the level, the smaller the image patch.

In Fig.~\ref{fig:fig4}b, we show examples of MIRCs generated from three different images for the zebra class from the three datasets and their deepest MIRCs that have the highest classification probabilities using the ERF227 and ERF11 models. 
These examples show that on the one hand, the ERF227 model was able to classify the image with high classification probability by relying exclusively on relatively local features, i.e. the zebra's face or stripes. 
On the other hand, the ERF11 model required larger image patches for successful classification, especially on the Sketchy dataset. 
This seems to indicate that the model with the larger ERF actually requires a much smaller part of the image to reach the correct classification as compared to the model with the smaller ERF. 

To verify whether this finding holds in general, we computed the histograms of the deepest MIRC levels for each image for all datasets and models (Fig.~\ref{fig:fig4}c-e).
We observed for the base models a dependence between ERF size and maximal MIRC levels, i.e., the larger the ERF size of the CNN, the higher its maximal MIRC levels (i.e. a smaller part of the image was sufficient to classify) (Fig.~\ref{fig:fig4}c). 
By contrast, networks with smaller ERFs typically cover a larger part of the image or the entire image for classification. 
We found this dependence to be dataset-specific.
The difference between ERF227 and ERF11 was largest for the Sketchy dataset and smallest for the Animals dataset.
The difference between ERF227 and models with smaller ERFs was reduced after adding the follow-up network without spatially scrambling the features (Fig.~\ref{fig:fig4}d). 
However, the difference was not affected when a follow-up network was added after spatially scrambling the features during training (Fig.~\ref{fig:fig4}e). 
The effect of feature-scrambling on the distribution of the levels of MIRCs demonstrates the different strategies CNNs can employ for object recognition.
On the one hand, spatial integration of features without scrambling led the follow-up networks to be able to construct and be selective to more reliable coarse-grained features than the base models.
Subsequently, these models (base + follow-up) had smaller MIRCs than their base models.
On the other hand, spatially scrambling the features before feeding them to the follow-up networks prevented them from exploiting the spatial relations between the features to construct more reliable coarse-grained features. 
Therefore, the follow-up networks were only able to learn the set of more fine-grained features that correlates with the target class.
Subsequently, these models retained the relatively large-sized MIRCs of their base models.

To visualize the features required for recognizing a certain class, we obtained latent representations for all  MIRCs of all images of a given class using  the model.
We then used the k-means algorithm to group the latent representations into 5 clusters.
In Fig. A.\ref{fig:mircclusters}, we show examples for the horse and eyeglasses classes of the Sketchy dataset for the model ERF227.
For each cluster, we show the eight MIRCs that are the closest to the cluster center and originate from distinct images.
We observe that each cluster is composed of MIRCs that represent visually similar features.
For example, we observe clusters representing hair, the side view of the head, and leg features for the horse class~\ref{fig:mircclusters}.
For the eyeglasses class, we can identify a cluster containing double-lined frames, one for thin frames, and one for reflective glass features.

\section{Discussion}
Despite the exceptional performance of CNNs in object recognition tasks~\cite{Krizhevsky2012, He2016}, the nature of their representations is still poorly understood. 
One aspect of the learned representations in CNNs is how they represent the shapes of objects and whether they are capable of encoding spatial relations among object parts.
It has recently been reported that CNN representations may be mostly local~\cite{Brendel2019, Baker2020} and consequently more biased toward object surface regularities~\cite{Jo2017MeasuringTT, Geirhos2019} than the global form of objects.
This led to the hypothesis that they might not be capable of representing spatial relationships among features~\cite{baker_deep_2022, Baker2018}.
In contrast to conclusions drawn in other works, our analysis allows us to provide the following more nuanced view: 
(1) We provide evidence that CNNs are capable of using relatively long-range spatial relationships for object classification, especially for textureless datasets (such as sketches). 
This finding is supported by several analyses, including a new scrambling approach in which we perturbed spatial relations between features within the CNN, and a systematic investigation of how CNN performance is impacted by different effective receptive field sizes.  
(2) We show that CNNs use different strategies for different datasets, rather than one unified strategy (e.g. pooling evidence based on local texture).  
Notably, we found that classification strategies can even vary even between classes within the same dataset. 
These strategies differ in the granularity of the features used and in the degree of reliance on the spatial relations between them. 
This suggests that there is a continuous spectrum of CNN strategies, ranging from exclusive reliance on local features (insensitive to spatial relations, found for example for the Animals dataset and the scrambling-insensitive classes in ImageNet) to a very strong reliance on spatial relations (for example for the Sketchy dataset and the scrambling-sensitive classes in ImageNet). 
(3) We furthermore show to what extent spatial relations among features are used by CNNs to perform object recognition tasks. 
In particular, we provide evidence that the spatial arrangement of features is used only to construct features up to an intermediate level of granularity. 
That is, we did not find evidence of spatial integration in CNNs that allows them to capture the global shape of the objects in the datasets tested. 

One possible explanation for a bias towards local features is the locality of the convolution operation~\cite{Baker2020}. 
However, our finding that CNNs learned features of intermediate granularity for classification agrees with another possible explanation, namely that a bias to local features is a consequence of the optimization process.
Specifically, from an information-theoretic perspective, features of intermediate granularity are the most informative for image classification tasks~\cite{Ullman2002}.
The idea is that, on the one hand, very complex features could be highly diagnostic because their presence gives high confidence about the class identity. 
However, on the other hand, these complex features may not be sufficiently sensitive (i.e. they do not exist in each exemplar) to be generalizable across exemplars of an object class. 
By contrast, very simple features would generalize better, but in addition would also lead to more false positives (i.e. lower specificity).
Thus, features of intermediate complexity can provide an optimal trade-off between sensitivity and specificity~\cite{Ullman2002}. 
A recent finding that agrees with this concept is that CNNs develop a bias in their representations towards the relational structure of the features only when these relational changes are diagnostic of certain object categories~\cite{malhotra_human_2022}. 
A similar concept is the idea of simplicity bias of neural networks, which states that neural networks preferentially extract the simplest features needed to solve a given task~\cite{shah2020pitfalls, malhotra2020hiding}. 
Consistent with this explanation, our MIRC analysis showed that models with small ERFs that by design are only capable of extracting simpler fine-grained features require larger patches of images for correct object recognition (because they have lower specificity).
In contrast, models with larger ERFs that are capable of extracting more coarse-grained and more specific features were able to assign objects to their corresponding correct classes based on smaller image patches. 
Therefore, our results suggest that optimization for object recognition is unlikely to yield bias to the global shape of objects, even if the models have the capacity to learn it.
A similar principle may hold for human vision, as it has been shown that in humans shape bias can be task- and context-dependent~\cite{diesendruck2003specific, cimpian2005absence, yoshida2003shifting}. 

Our results have major implications for the ongoing discussion concerning shape and texture representation in CNNs, and whether certain biases exist.
There is little consensus about the extent to which CNNs are texture- or shape-biased. 
Some studies have suggested that CNNs are shape-biased~\cite{cognitive_shape_2017, tartaglini_developmentally-inspired_nodate, kubilius_deep_2016}, whereas others have suggested that CNNs are strongly texture-biased~\cite{Geirhos2019, baker_deep_2022, Baker2018, Baker2020}. 
Here, instead of using a shape-texture dichotomy to understand the nature of CNN representations, we have used the dichotomy of local vs. global features. 
We argue that this dichotomy is useful for two reasons: 
1) It can be quantified without specific interpretations of what constitutes texture or shape, as we showed with our feature-scrambling approach. 
In fact, our approach is hypothesis-free to some extent because we do not perform specific image manipulations to provide evidence for either texture or shape bias. 
Rather, we manipulate the network architecture and the spatial arrangement of the representations to determine the locality of the features. 
2) It is flexible in that it allows local features to be both shape-like or texture-like. 
This means that the shape-texture dichotomy only maps partially to the global-local dichotomy.
For example, this dichotomy is able to account for the existence of highly diagnostic shape features of fine granularity that are highly specific and sensitive (e.g., the nose of a dog). 
Indeed, when ranking ImageNet classes according to their scrambling sensitivity, it is not always obvious that the scrambling-sensitive classes would map to shape classes as would be intuitively expected. 
A possible explanation for previous inconsistent findings with respect to shape and texture is that the respective studies made very specific manipulations that did not generalize beyond these examples. 
For example, the texture bias observed in CNNs trained on ImageNet when tested on shape-texture cue conflict stimuli~\cite{Geirhos2019} was significantly reduced when the background of the images was removed~\cite{tartaglini_developmentally-inspired_nodate}.
Our findings suggest an explanation for these observations, in that the fine-grained (texture) features are less reliable than the more coarse-grained (shape) features, and therefore need to cover a large portion of the image to be diagnostic. 
Removing them from the background reduced their predictive power and led CNNs to be more shape-biased~\cite{tartaglini_developmentally-inspired_nodate} (on this specific test set).
Another example is that many studies used silhouette stimuli to test shape bias in CNNs~\cite{Geirhos2019, baker_deep_2022, kubilius_deep_2016} and reached different conclusions.
However, they used different datasets containing different classes.
According to our results, this is expected since CNNs employ different classification strategies per object class and consequently will lead to variable classification performances on silhouette stimuli if the classes are different.

Given that CNN models are currently used as models of brain activity, specifically for the ventral stream of the visual system, which is believed to be responsible for object recognition~\cite{Yamins2016, Yamins2014a, Cadieu2014a, cichy_comparison_2016}, it is important to understand the representations they develop and how they deviate from the brain.
Recent evidence has shown that the categorical organization of the entire ventral stream can be explained by mid-level features that do not include intact objects and do not convey any semantic information~\cite{Long2018, jagadeesh2022texture, henderson2022texture}. 
Moreover, it has been reported that human children's ability to recognize objects based on their global shape begins to develop only at 18-24 months of age~\cite{Pereira2009}. 
Before that, they are capable of recognizing objects based solely on their local features. 
Furthermore, humans in general are capable of recognizing familiar objects from local image patches~\cite{Ullman2016} and these image patches indeed evoke responses in higher-order category-selective visual areas~\cite{holzinger_minimal_2019}.
Our results, therefore, provide additional evidence for the hypothesis that features of intermediate granularity are optimal for image classification~\cite{Ullman2002, Ullman2001}.
In summary, we showed here that although CNNs do not exploit global shape representations to perform object recognition, they can learn to utilize distributed feature constellations if this is required for solving the object classification task at hand.

Looking ahead, we hypothesize that developing new tasks and objective functions to train CNNs, such as communicative efficiency~\cite{portelance2021emergence} or action planning using reinforcement learning~\cite{lindsay2021divergent} can not only lead to more effective biases and representations in such networks but also shed more light on how the observed human biases emerge.

\section{Conclusions}
We provide evidence that CNNs have the capacity to learn the spatial relations between features for object recognition. 
Specifically, the spatial arrangement of features is exploited by CNNs to build more coarse-grained features that are more reliable for object classification.
Notably, the capacity of CNNs to learn the spatial arrangement of features varies according to the dataset and according to the class within the same dataset.
We noticed, however, that CNNs employ the spatial configuration of features to build more coarse-grained features only up to an intermediate degree of granularity and do not exploit the global shape of objects. 
The reason for this is that features of intermediate granularity are more likely to be optimal in the trade-off between sensitivity and specificity i.e. generalizable and yet reliable.   
\section*{Declaration of competing interest}
The authors declare that they have no known competing financial interests or personal relationships that could have appeared to influence the work reported in this paper.
\section*{Acknowledgments}
This research did not receive any specific grant from funding agencies in the public, commercial, or not-for-profit sectors.

\bibliographystyle{elsarticle-num} 
\bibliography{sample}
\clearpage
\newpage
\section*{Appendix}
\textbf{Controlling for the number of parameters of the models.} 
We performed a control analysis to verify that the performance differences observed in our study among CNNs of different ERFs can be attributed indeed to their ERFs and not the number of model parameters.  
We trained a wider model of small ERF ($11\times11$ pixels) but \emph{with} matched the number of parameters to the model with the largest ERF ($227\times227$ pixels). 
For both the Animals and Sketchy datasets, we observed a slight increase in the classification performance of the models by increasing the number of parameters. 
However, a small ERF model with a large number of parameters did not reach the performance of the model with the largest ERF, indicating the importance of the ERF to the models' performance.
Furthermore, for the Sketchy dataset, the performance of the wider model with ERF = $11\times11$ did not even reach the performance of the regular model with ERF = $15\times15$ pixels. 
This is in line with our other results showing the reliance of the performance of CNNs on their ERF size, especially for the Sketchy dataset. 
Note that in the manuscript, we included several additional controls, e.g. scrambling during training, a $1\times1$ follow-up network, and local scrambling, which further show the importance of ERF size. 

\begin{suppfigure*}[h]
\includegraphics[width=\linewidth]{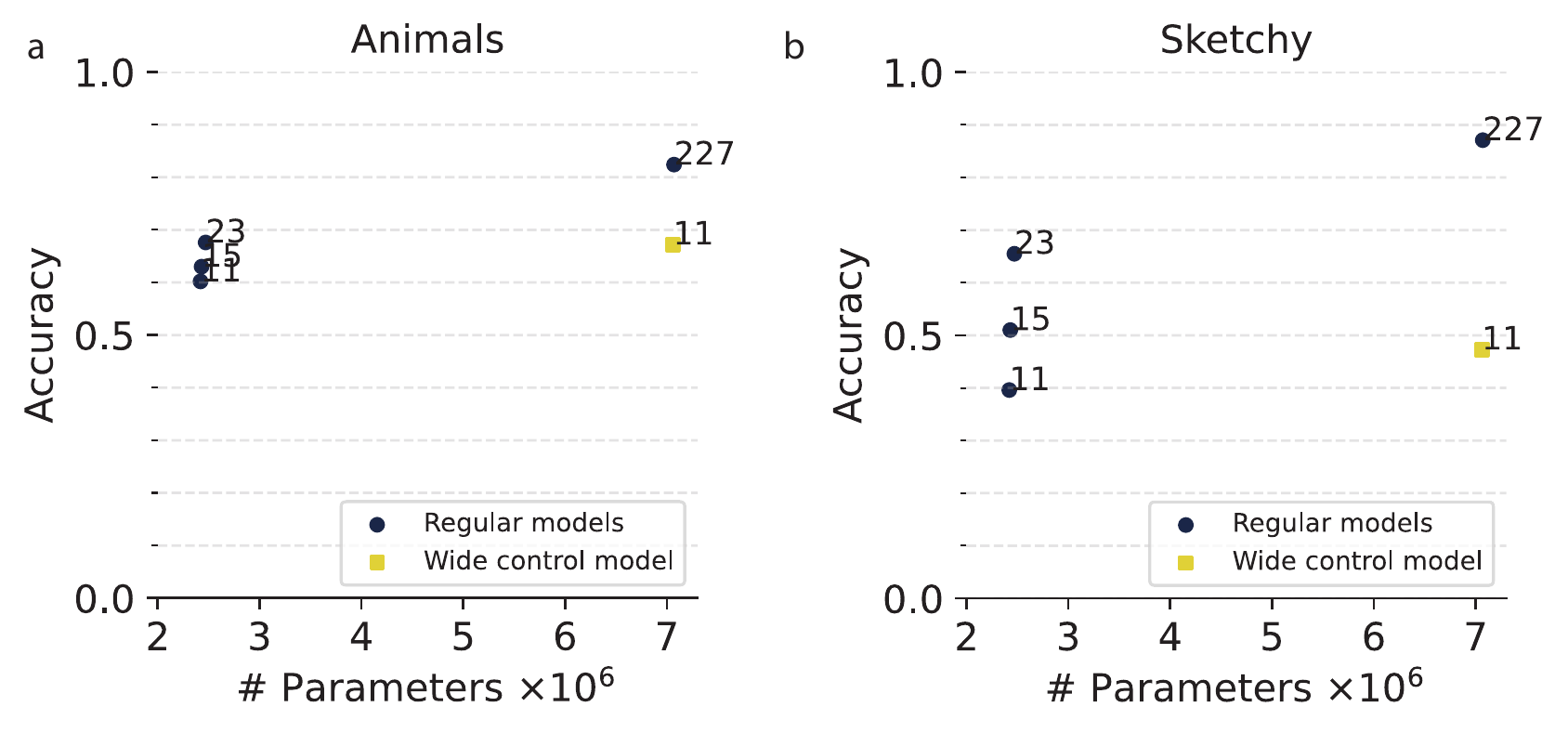}
\caption{
A control experiment in which we trained a wide model that has a small ERF (11 pixels), while matching the number of parameters of the model with the largest ERF (227 pixels). 
The numbers shown in the figure are the ERF of the corresponding models in pixels.
}
\label{fig:control}
\end{suppfigure*}
\clearpage

\begin{suppfigure*}[h]
\includegraphics[width=\linewidth]{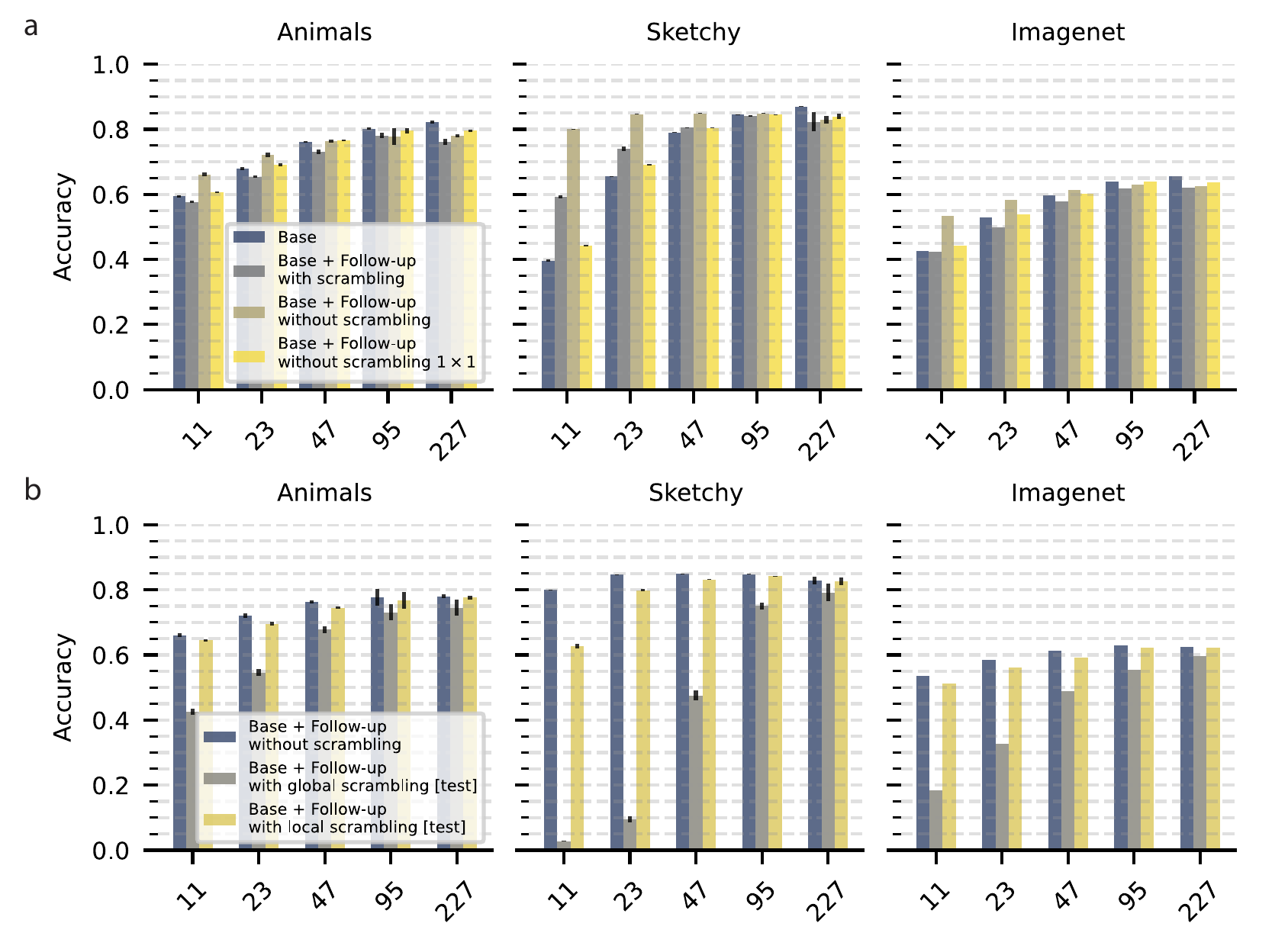}
\caption{
\textbf{(a)}: Classification accuracy for CNN models of different ERFs under different training conditions of the feature-scrambling approach (Fig.~\ref{fig:fig1}c).
\textbf{(b)}: Classification accuracy of the base models with spatial aggregation without scrambling under different testing conditions (global and local scrambling).
}
\label{fig:rawdata}
\end{suppfigure*}

\clearpage

\begin{suppfigure*}[h]
\includegraphics[width=\linewidth]{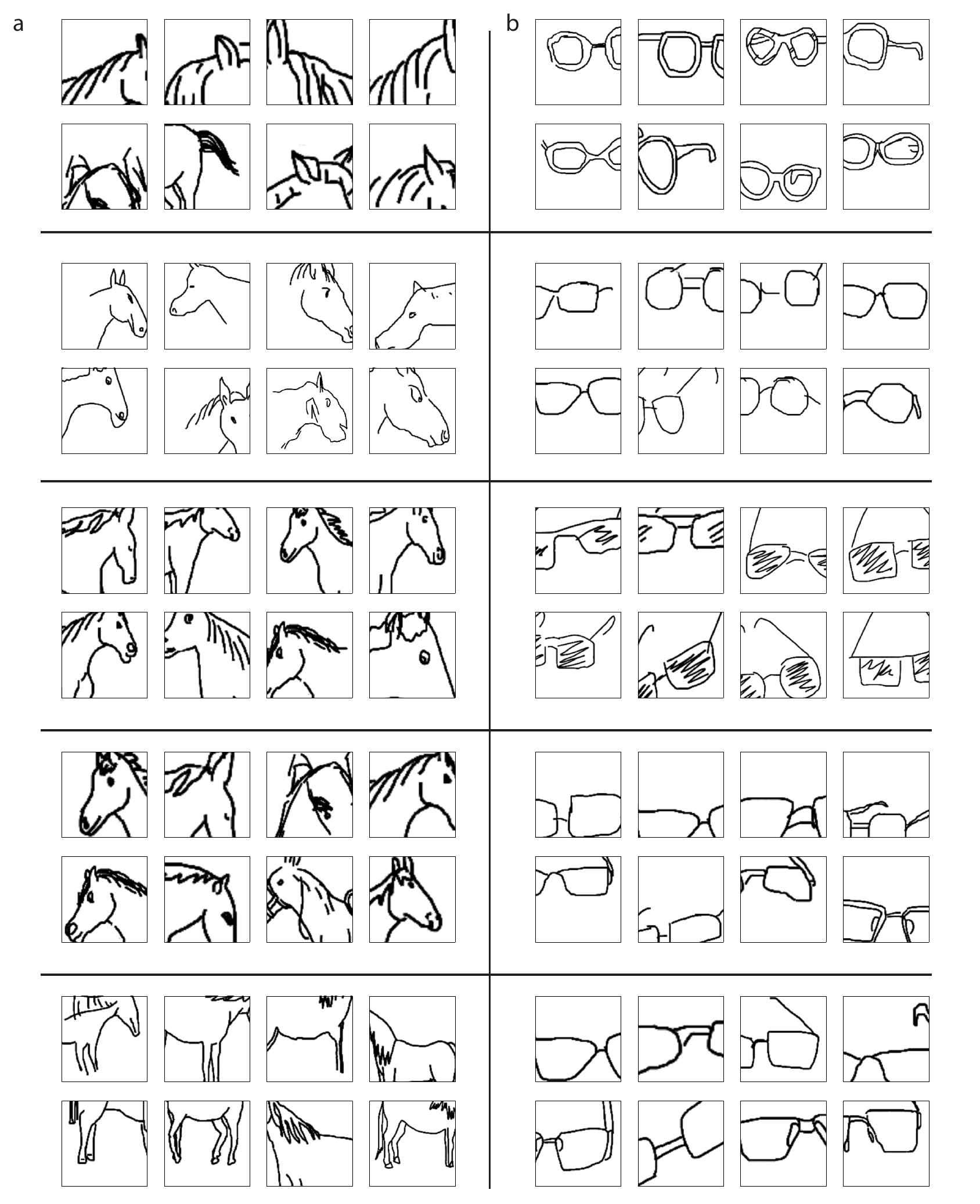}
\caption{
\textbf{Clustering of all the MIRCs of the horse (a) and eyeglasses (b) classes (Sketchy dataset) in the representational space of the model ERF227}
Each panel shows the eight closest MIRCs, generated from unique test images, in the representational space to the center of one cluster. 
}
\label{fig:mircclusters}
\end{suppfigure*}

\end{document}